\documentclass[conference]{IEEEtran}
\IEEEoverridecommandlockouts
\usepackage{cite}
\usepackage{amsmath,amssymb,amsfonts}
\usepackage{algorithmic}
\usepackage{graphicx}
\usepackage{textcomp}
\usepackage{xspace}
\usepackage[dvipsnames]{xcolor}
\usepackage{booktabs}
\usepackage{color,soul}
\usepackage{hyperref}

\usepackage{fancyhdr}

\fancypagestyle{firstpage}{
  \fancyhf{} 
  
  \fancyhead[L]{\small{$\copyright$ 2025 IEEE. Personal use of this material is permitted. Permission from IEEE must be obtained for all other uses, in any current or future media, including reprinting/republishing this material for advertising or promotional purposes, creating new collective works, for resale or redistribution to servers or lists, or reuse of any copyrighted component of this work in other works}}
  \fancyfoot[C]{\thepage}
}

\usepackage{todonotes}

\newcommand{\ie}{i.\,e.,\xspace}

\def\BibTeX{{\rm B\kern-.05em{\sc i\kern-.025em b}\kern-.08em
    T\kern-.1667em\lower.7ex\hbox{E}\kern-.125emX}}

\title{Integrating Prior Observations for Incremental 3D Scene Graph Prediction\\
}
\author{Marian Renz$^{1}$, Felix Igelbrink$^{1}$, Martin Atzmueller$^{1,2}$
\thanks{The DFKI Niedersachsen (DFKI NI) is sponsored by the Ministry of Science and Culture of Lower Saxony and the VolkswagenStiftung. This work is supported by the Federal Ministry of Research, Technology and Space (BMFTR) within the ExPrIS project (grant number: 01IW23001)}
\thanks{$^{1}$Cooperative and Autonomous Systems, DFKI Niedersachsen, German Re\-search Center for Artificial Intelligence, Osnabrück, Germany\newline{\tt\small \{firstname\}.\{lastname\}@dfki.de}}%
\thanks{$^{2}$Semantic Information Systems, Osnabrück University, Osnabrück, Germany}

}

\begin{document}


\maketitle

\begin{abstract}
3D semantic scene graphs (3DSSG) provide compact structured representations of environments by explicitly modeling objects, attributes, and relationships. While 3DSSGs have shown promise in robotics and embodied AI, many existing methods rely mainly on sensor data, not integrating further information from semantically rich environments. Additionally, most methods assume access to complete scene reconstructions, limiting their applicability in real-world, incremental settings. This paper introduces a novel heterogeneous graph model for incremental 3DSSG prediction that integrates additional, multi-modal information, such as prior observations, directly into the message-passing process. Utilizing multiple layers, the model flexibly incorporates global and local scene representations without requiring specialized modules or full scene reconstructions. We evaluate our approach on the 3DSSG dataset, showing that GNNs enriched with multi-modal information such as semantic embeddings (e.g., CLIP) and prior observations offer a scalable and generalizable solution for complex, real-world environments. The full source code of the presented architecture will be made available at \url{https://github.com/m4renz/incremental-scene-graph-prediction}.
\end{abstract}


\thispagestyle{firstpage} 

\section{Introduction}

Semantic scene graphs (SSGs) offer a structured and compact representation of visual environments by explicitly modeling objects, their attributes, and inter-object relationships in a semantically rich way. Initially developed for 2D image understanding, the extension of SSGs into the 3D domain \cite{Armeni2019-ip} has gained significant traction, particularly in robotics, where spatial reasoning and situational awareness are critical. 3D semantic scene graphs (3DSSGs) enable environmental modeling by incorporating geometric and topological information. In consequence, this allows for a more accurate interpretation of complex scenes.

Furthermore, 3DSSGs serve as a powerful bridge between raw sensory input and high-level semantic understanding by facilitating the integration of multi-modal information, such as additional sensory data and even common-sense knowledge. As a result, they are increasingly adopted in robotics research as a foundational representation for embodied AI systems that require both perceptual grounding and semantic reasoning.
The construction or generation of 3DSSG from sensor data has therefore become a prominent topic in machine learning and robotics \cite{Hughes2022-fg, Bavle2023-iz, Looper2023-kb}.

With progress in 3DSSG inference using Graph Neural Networks (GNNs), a variety of approaches have emerged that integrate 3DSSG generation with additional information sources. These methods leverage supplementary data to refine object and relationship predictions, enhance generalization across environments, and support downstream tasks such as navigation \cite{Ravichandran2022-vd}, exploration \cite{Li2022-ni}, or task planning \cite{Agia2022-mi}. However, while all integration approaches independently show promising results, they lack a generalized mechanism that is agnostic to the modality of the utilized information, making them highly dependent on the utilized training data.

Moreover, most existing approaches focus on inferring 3DSSGs from fully reconstructed scenes, where complete geometric information is available at inference time \cite{Wald2020-yj}. This makes these approaches impractical for many real-world tasks, where a scene is typically captured incrementally from a stream of sensor data. Incremental SSG generation requires models to utilize information acquired from prior observations to predict and interpret new sensor inputs.

In this work, we present a method for incremental 3DSSG generation by integrating the sub-tasks required for the SSG construction into a multi-layered architecture. This design allows for flexible incorporation of multi-modal information into the model architecture without the need for specialized modules. This is not only limited to new features, it also extends to topologically different graphs. 

Central to our approach is a heterogeneous scene graph design that fuses sensor data and observations from previous time steps across \emph{global} and \emph{local layers}. Global layers provide spatial, geometric, and semantic context for the entire scene, while local layers integrate current sensor data. The proposed model efficiently stores and integrates spatial, geometric, and semantic features by embedding them directly into the message-passing process, eliminating the need to store numerous point-cloud segments or time-series data.


The main contributions of our work are summarized as follows:

\begin{itemize}
    \item We propose a novel heterogeneous graph model for 3D scene graph generation that integrates multi-modal information for incremental prediction.
    \item We evaluate the proposed model on the 3DSSG dataset for per-frame incremental scene graph prediction.
    \item We demonstrate the robustness of our model against erroneous predictions in prior observations.
\end{itemize}




\section{Related Work}


SceneGraphFusion \cite{Wu2021-hw} was the first approach to generate scene graphs incrementally. The authors infer local 3D scene graphs from partial point cloud segments derived from individual RGB-D frames in the 3DSSG dataset \cite{Wald2020-yj}. These local scene graphs are subsequently fused into a global graph. A variation of this method has also been applied using only RGB image sequences \cite{Wu2023-dm}. However, the proposed model utilizes only the updated geometrical information when predicting novel frames. The existing global scene graph, generated from previous frames, remains invisible to the model. As a result, the model does not benefit from prior knowledge of the scene structure and instead predicts each frame independently.
In our approach, we directly integrate prior predictions by linking instances from frames to previously predicted nodes, thus enabling our model to benefit from earlier observations without the need to store the fully segmented point cloud.

Most similar to our idea, Feng et al. \cite{Feng2025-cz} incorporate historical predictions for incremental scene graph generation, using a recurrent mechanism to integrate the last \textit{m} processed graphs and embedding a global graph representation as a one-hot encoded matrix into the prediction process. In contrast, we do not encode global information explicitly; instead, we integrate it directly into the message passing by linking past predictions and matching node instances.  This approach allows newly integrated information to directly enhance downstream SSG prediction. Furthermore, we explore the use of heterogeneous GNNs to improve the integration of semantically relevant information.

In the context of multi-modal integration, several recent approaches have investigated heterogeneous graph structures and external knowledge sources. Ma et al. \cite{Ma2024-gd} infer relationship types based on three top-level categories from the 3DSSG dataset and apply heterogeneous message passing on the learned graph structure. Directed Spatial Commonsense Graphs (D-SCG) \cite{Giuliari2023-sw} incorporate heterogeneous information from ConceptNet \cite{Speer2017-os} with 3DSSGs to localize objects in partial 3D scenes. Knowledge-Scene Graph Networks \cite{Qiu2023-ou} integrate external knowledge curated from multiple sources using GB-Net \cite{Zareian2020-kc}, embedding this knowledge directly into the message passing process. While these approaches successfully integrate multi-modal information for their respective tasks, none have been applied to the problem of incremental 3DSSG generation.

\section{Method}
\label{sec:method}

\subsection{Dataset and Preprocessing}
\label{chap:dataset}

To train and evaluate our proposed method, we utilize the 3DSSG dataset, which extends the 3RScan dataset \cite{Wald2019-ki} with scene graph annotations for over one thousand indoor 3D scenes created using RGB-D reconstruction.

We use the RIO27 label set, which features a total of 27 object categories and 16 relationship categories derived from \cite{Zhang2021-tl}. After filtering out invalid scenes, we obtained a total of 1,320 usable 3D scenes from the dataset. 
Each scene consists of an annotated reconstruction of the geometry, available as a 3D mesh, the complete scene graph, as well as the raw RGB-D frames and their poses used to reconstruct the scene.

Since, for incremental scene graph prediction, neither the full scene geometry nor the complete graph is available to the model at inference time, we extracted for every RGB-D frame $F_t$ the visible geometry as a point cloud with instance annotations, along with the currently visible portion of the ground-truth scene graph. Additionally, for each frame $F_t$, we included the partial scene graph constructed from the preceding frames $\{F_{0}, \dots, F_{t-1}\}$ (see Section~\ref{chap:graph_model}).

\subsection{Graph Model}
\label{chap:graph_model}
The core intuition behind our heterogeneous modeling approach is to connect previously observed objects in the sensor data stream to the same objects in newly recorded frames. This enables the model to leverage information from earlier observations during prediction.

When represented as a scene graph, this results in a two-layer architecture: a global scene graph that accumulates observations, objects, and relationships from previous frames, and a local scene graph constructed from the sensor data of a single frame. The target task is to predict the classes and relationships in the local scene graph by utilizing both the sensor data and the global scene graph.

Assuming a partial global scene graph with already classified objects and predicates, as provided by the dataset preprocessing (see Section~\ref{chap:dataset}), the first step is to perform object segmentation on the current frame to identify visible object instances for the local graph. For this work, we use the ground-truth segments from the 3DSSG dataset. To construct the local graph, we convert the depth frame into a point cloud and add bidirectional edges between objects that are less than 0.5 meters apart, following the approach in \cite{Wu2021-hw}. During training, we also use ground-truth information to match local nodes to previously seen nodes in the global graph.

\begin{figure}[htb]
    \centering
    \includegraphics[width=\linewidth]{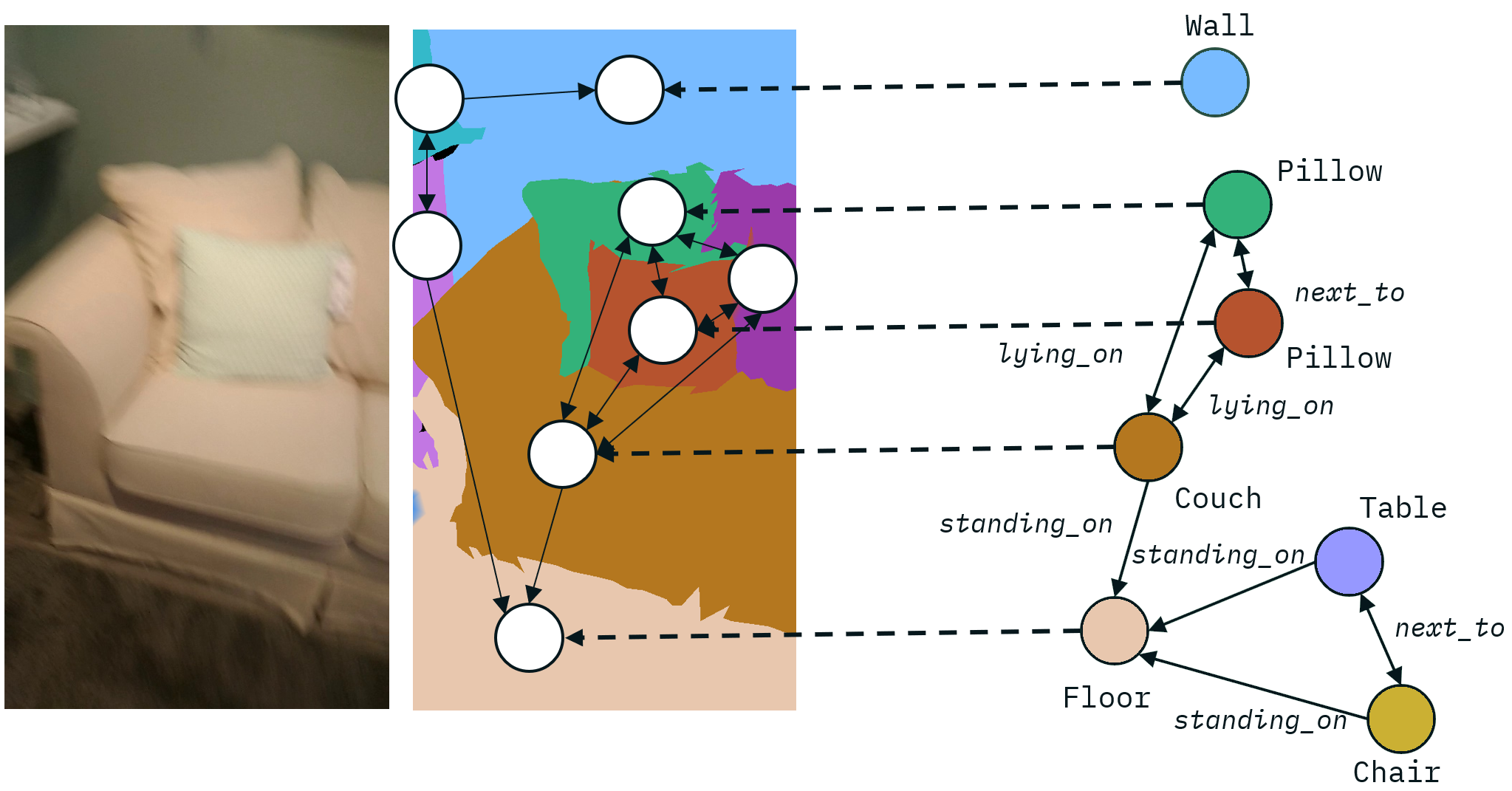}
    \caption{Example of the heterogeneous scene graph. The left image shows the RGB-D frame from \cite{Wald2020-yj}, the middle shows the segmented frame with the unpredicted local scene graph, and the right shows the already predicted global scene graph. Dashed lines represent the edges between matched nodes.}
    \label{fig:hetero_sg}
\end{figure}

Nodes and edges in the global and local scene graphs are modeled as distinct node and edge types within a single heterogeneous graph. Additional edges connect global and local nodes for all matched node pairs (see Fig.~\ref{fig:hetero_sg}), allowing information to flow from the global to the local graph.

The node features are similar to those used in \cite{Wu2021-hw}. For each object, 256 points are sampled from the point cloud. Additionally, a hand-crafted descriptor is computed, consisting of the center $c$ and standard deviation $std$ of the sampled points, bounding box side lengths $l$, $w$, and $h$, the maximum bounding box length $L$, and the bounding box volume $V$. Global nodes also include information from previous predictions, either as a class label or as a CLIP \cite{radford2021learning} embedding of the predicted label.

To merge the local scene graph into the global one, the points of matched nodes are downsampled again to 256 points, and the descriptor is recalculated. Ground-truth labels and instance identifiers remain unchanged. New nodes and edges are added directly to the global graph. 

\subsection{Graph Neural Network Architecture}

\begin{figure*}[htb]
    \centering
    \includegraphics[width=\textwidth]{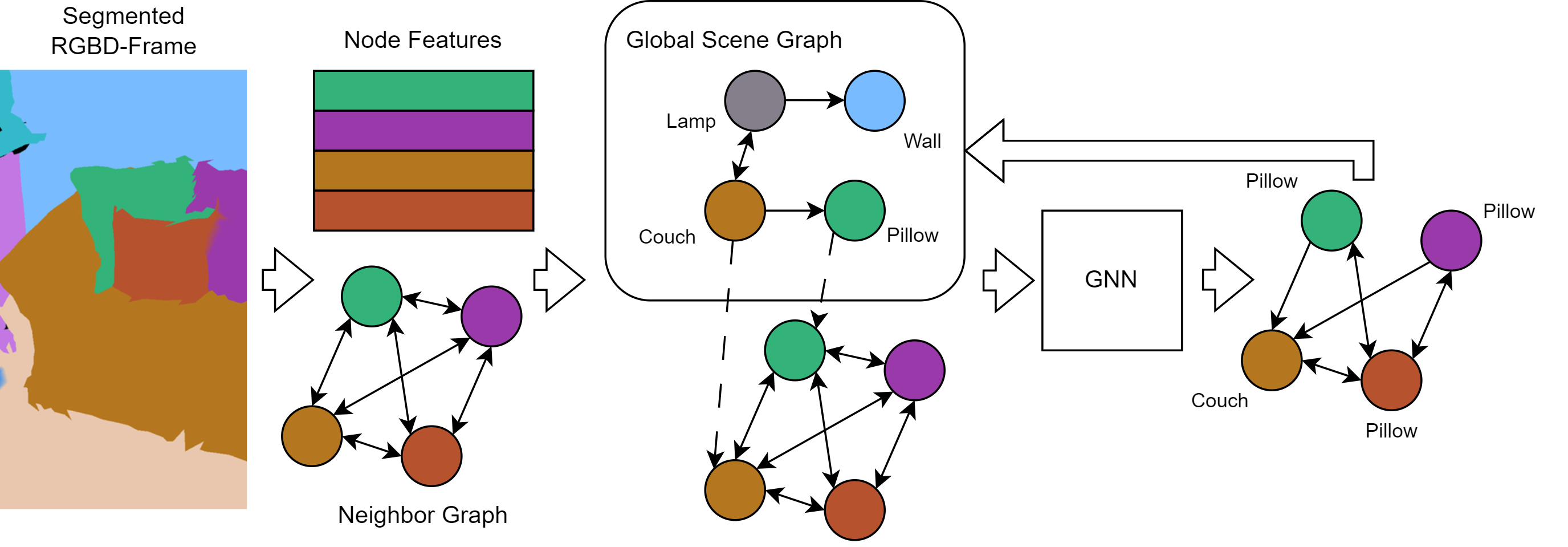}
    \caption{Pipeline of the proposed approach. At timestep $t$, a local frame graph is constructed from a segmented RGB-D frame based on a neighborhood graph and segment-specific node features. The local graph is then connected to a globally constructed scene graph from frame $t-1$. After message passing, node and edge classes are predicted, and the local graph is merged into the global graph.}
    \label{fig:architecture_1}
\end{figure*}

\begin{figure}[htb]
    \centering
    \includegraphics[width=\linewidth]{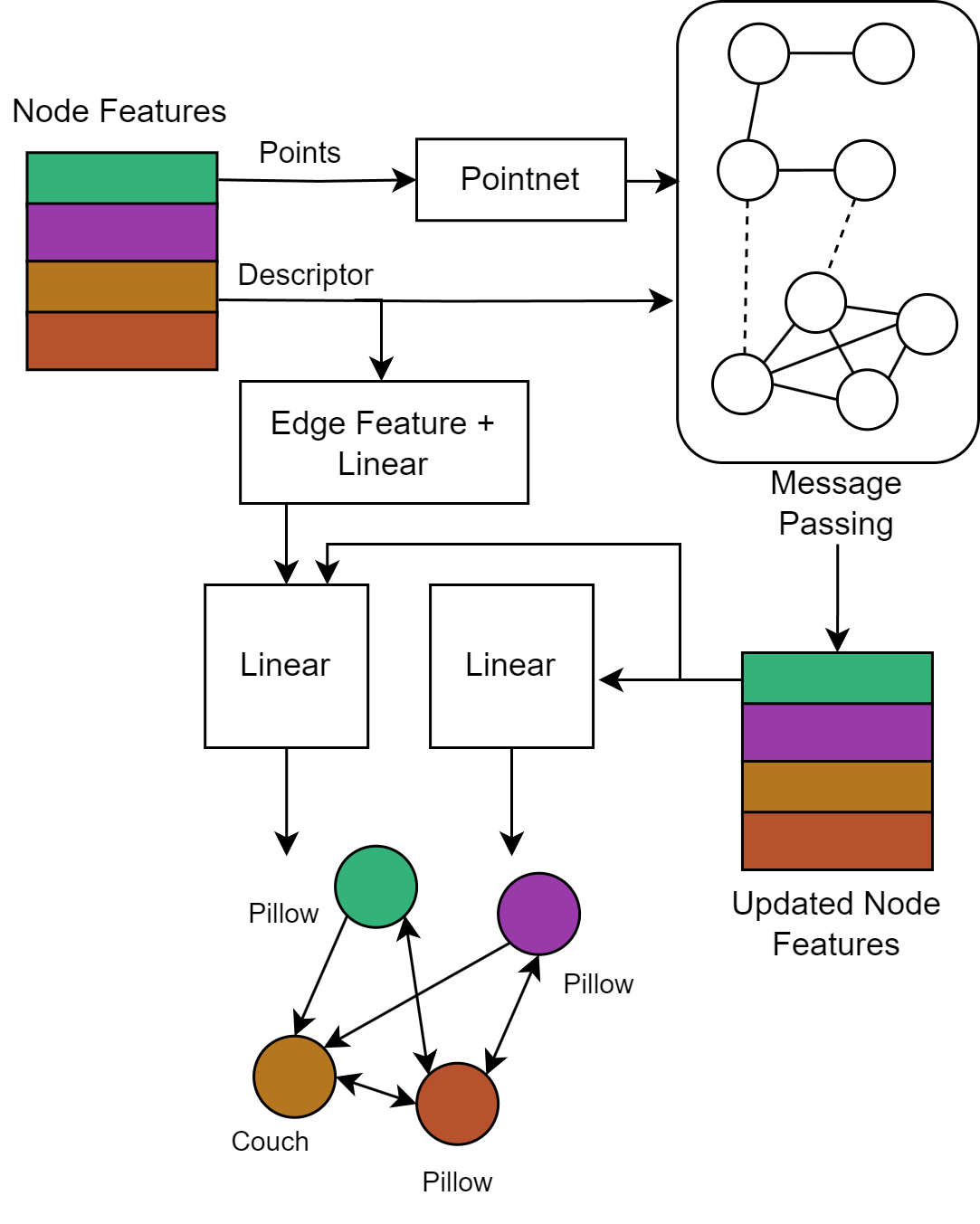}
    \caption{GNN architecture for 3D scene graph generation. For node features, the sampled points are passed through a PointNet encoder and concatenated with a geometric descriptor. For edge features, the descriptor for an edge $e_{ij}$ connecting nodes $i$ and $j$ is formed by concatenating descriptors $d_i$ and $d_j$, which are then passed through two linear layers. After message passing, node classes are predicted using two linear layers applied to the updated node features. For edge classification, the edge feature of $e_{ij}$ is concatenated with the updated node features of nodes $i$ and $j$, and passed through two linear layers as well.}
    \label{fig:architecture_2}
\end{figure}

Each node's feature representation is constructed by embedding 256 sampled points using a PointNet encoder \cite{Qi2017-wl}, and concatenating the resulting point feature with a geometric descriptor vector. For global nodes, an additional feature vector is included based on the ground truth label, either as a one-hot encoded class index or a CLIP embedding of the text label. Edge features are computed by subtracting centroid and standard deviation offsets, and applying logarithmic ratios of bounding box dimensions, following the approach of \cite{Wu2021-hw}. Specifically, the edge feature vector for an edge $e_{ij}$ between nodes $i$ and $j$ is defined as:
\[
[c_j - c_i, \; \text{std}_j - \text{std}_i, \; \log(\tfrac{l_j}{l_i}, \tfrac{w_j}{w_i}, \tfrac{h_j}{h_i}), \; \log(\tfrac{L_j}{L_i}), \; \log(\tfrac{V_j}{V_i})],
\]
and is passed through a two-layer MLP. Message passing is performed using a two-layer GNN, either a heterogeneous GraphSAGE \cite{Hamilton2017-nc, Schlichtkrull2018-fx} or HGT \cite{Hu2020-gs}. Node classification is conducted via a two-layer MLP applied to the updated node features. For edge classification, the updated source and target node features are concatenated with the edge feature vector. Note that edge features are not used during message passing due to limitations of the employed GNN layers. All layers use ReLU activation, layer normalization, and a dropout rate of $0.5$, except for the final layers of each sub-network.

As baselines, we include homogeneous GraphSAGE and SGFN \cite{Wu2021-hw} applied only to the local frame graph. Additionally, we evaluate a homogeneous version of the global-local heterogeneous graph, where missing label features in local nodes are replaced with $-1$, and missing CLIP embeddings with zero vectors to ensure consistent feature dimensions. Edges between global and local nodes in this setting do not carry ground truth labels. We also test a variant of the heterogeneous architecture without ground truth features in the global layer. To assess robustness, we additionally train the homogeneous SAGE and HGT/heterogeneous SAGE with 20 $\%$ and 50 $\%$ falsified global labels, resulting in incorrect CLIP embeddings.

\subsection{Training Details}

The model is trained using a composite loss function that combines a weighted node classification loss and a weighted binary edge classification loss, scaled by a factor $\alpha=40$ for positive edge classes. The total loss is defined as
\begin{equation}
    \mathcal{L} = w_n\mathcal{L}_n + \alpha w_e \mathcal{L}_e,
\end{equation}
where the weights are computed based on the inverse log-frequency of class occurrences, \ie
\begin{equation}
    w_n = \frac{10}{\log(n_n)}, \quad w_e = \frac{10}{\log(n_e)} + 1.
\end{equation}

Only predictions for local nodes and edges are considered during loss and gradient computation across all models. Note that node classification is treated as a multi-class problem, with only one correct ground truth class per object, whereas edge prediction is a multi-label task, allowing multiple valid ground truth labels per edge.

Training is conducted for up to 100 epochs, with early stopping triggered if the validation loss does not improve for 5 consecutive epochs. For SAGE models, a learning rate of 0.0001498 is used with a step scheduler (decay factor $\gamma = 0.05$, step size = 30), while HGT models are trained with a learning rate of 0.0001 and a step scheduler with $\gamma = 0.05$ and a step size of 20. The SGFN model follows the training procedure of the GraphSAGE model, using the hyperparameters and loss function described in \cite{Wu2021-hw}.


    

    

\section{Evaluation}

We train all models on the preprocessed dataset described in Chapter~\ref{chap:dataset}, following a 0.8/0.1/0.1 split for training, validation, and test data. For all models, we predict and evaluate only nodes and edges within the local frame, and not in the global graph.  Additionally, we evaluate models trained on data with 20 $\%$ and 50 $\%$ falsified ground truth labels in the global layer to assess the robustness of the approach.


\subsection{Metrics}

We evaluate four aspects of incremental scene graph prediction:  
(1) Node classification, measured by mean accuracy across all local nodes; 
(2) Edge classification, evaluated using mean recall;  (3) Relationship prediction, measured by the number of correctly predicted ground truth triples using the ng-Recall@k metric \cite{Gkanatsios2019-eo}, which determines the fraction of detected ground truth triples among the top-$k$ predicted triples of the local scene graph; and (4) Node classification for previously unseen nodes, i.e., nodes appearing for the first time in the sequence and not yet present in the global graph, also evaluated using mean accuracy.

\subsection{Scene Graph Prediction}


\begin{table*}[t]
    \centering
    \caption{Results for the incremental 3D scene graph generation. The first section shows results for homogeneous GNNs, the second section for heterogeneous GNNs and the third section for heterogeneous GNNs with additional edges.}
    \begin{tabular}{lccccccc}
    \toprule
        & Acc@1 & Acc@5 & Rec & ng-R@50 & ng-R@100 & U-Acc@1 & U-Acc@5 \\
    \midrule 
        SGFN & 0.48 & 0.80 & 0.31 & 0.00 & 0.00 & \textbf{0.40} & \textbf{0.83} \\
        SAGE & 0.28 & 0.65 & 0.64 & 0.00 & 0.00 & 0.37 & 0.80 \\ 
        SAGE+label & 0.92 & 0.98 & \textbf{0.81} & 0.02 & 0.01 & 0.30 & 0.70 \\
        SAGE+clip & \textbf{0.98} & \textbf{0.99} & 0.80 & 0.14 & 0.18 & 0.36 & 0.79 \\
    \midrule
        HetSage-plain & 0.34 & 0.76 & 0.6 & 0.07 & 0.09 & 0.31 & 0.71 \\
        HGT-plain &	0.42 & 0.81 & 0.63 & 0.16 & 0.20 & 0.21 & 0.58 \\
        HetSAGE+label & 0.73 & 0.98 & 0.74 & 0.53 & 0.61 & 0.31 & 0.70 \\
        HGT+label & 0.95 & 0.95 & 0.69 & 0.71 & 0.78 & 0.24 & 0.62 \\
        HetSAGE+clip & \textbf{0.98} & \textbf{0.99} & 0.75 & 0.58 & 0.69 & 0.33 & 0.75\\
        HGT+clip & \textbf{0.98} & \textbf{0.99} & 0.77 & \textbf{0.80} & \textbf{0.84} & 0.29 & 0.73\\
    \midrule
        HetSAGE+clip+add & \textbf{0.98} & \textbf{0.99} & 0.76 & 0.68 & 0.76 & 0.32 & 0.76 \\
        HetSAGE+clip+add-only & \textbf{0.98} & \textbf{0.99} & 0.73 & 0.51 & 0.61 & 0.34 & 0.78 \\
    \bottomrule
    \end{tabular}
    \label{tab:results}
\end{table*}

Our results (see Table~\ref{tab:results}) highlight several key findings. Models that operate solely on individual frames, such as SGFN and GraphSAGE, achieve the highest accuracy on previously unseen nodes, indicating that the introduction of prior classifications can impair generalization in node classification. However, these models perform poorly in predicting the overall scene graph structure. 
Incorporating a heterogeneous scene graph without additional semantic features (HetSAGE-plain and HGT-plain) yields only minor improvement on the relationship prediction, suggesting that structural information alone is insufficient. 
In contrast, enriching node features with simple labels or CLIP embeddings leads to consistently better performance, where clip embeddings lead to better results in all cases.
Homogeneous models perform equally well or better than heterogeneous models on node and edge prediction metrics, while heterogeneous models excel in relationship prediction. Among these, the HGT+label/CLIP models achieve the highest relationship prediction performance, with HGT+CLIP also showing competitive results across other metrics.  These results suggest that heterogeneous models are well-suited for capturing the rich semantic structures present in 3DSSGs.



\subsection{Robustness Against False Labels}

When introducing falsified labels into the global scene graph, we observe a general decline in performance across all metrics, except for mean edge recall, which slightly improves under 20 $\%$ falsified labels for all models except HGT. The most pronounced drop occurs in relationship prediction, with reductions ranging from $0.06$ to $0.34$ for 20 $\%$ falsified labels and from $0.11$ to $0.46$ for 50 $\%$ falsified labels at $k=50$ and $0.07$ to $0.37$ for 20 $\%$ falsified labels and from $0.13$ to $0.52$ for 50 $\%$ falsified labels at $k=100$.
Interestingly, the accuracy on previously unseen nodes is less affected, showing only a modest decline between $0.02$ and $0.07$, suggesting that the models still effectively learn to generalize prior classification features. Similar to the evaluation with correct labels, the homogeneous GNN with integrated prior observations outperforms its heterogeneous counterparts on node and edge prediction metrics. In contrast, heterogeneous models achieve better performance in relationship prediction. However, they also show the largest performance drop in this category when label noise is introduced.
Despite 20 $\%$ falsified labels representing a substantial level of noise, the relatively minor performance degradation in node and edge classification tasks across all models indicates a degree of robustness. However, the drop for relationship prediction suggests a strong reliance on prior observations, which has to be mitigated for real application scenarios. For most metrics (excluding unseen node accuracy), models trained with 20 $\%$ falsified labels consistently outperform those trained with 50\,$\%$, reinforcing the notion that the learned prior features remain informative even under moderate label corruption.

\begin{table*}[t]
    \centering
    \caption{Results for the 3D scene graph prediction with falsified feature classes in the global layer. The first section shows models trained with 20 $\%$ randomly falsified labels, the second section shows models trained with 50 $\%$ randomly falsified labels.}
    \begin{tabular}{lccccccc}
    \toprule
          & Acc@1 & Acc@5 & Rec & ng-R@50 & ng-R@100 & U-Acc@1 & U-Acc@5 \\
             \midrule 
        SAGE+clip+0.2 & \textbf{0.94} & \textbf{0.99} & \textbf{0.82} & 0.08 & 0.11 & \textbf{0.34} & \textbf{0.78}\\
        HetSAGE+clip+0.2 & 0.91 & 0.98 & 0.76 & 0.33 & 0.38 & 0.29 & 0.72\\
        HGT+clip+0.2 & 0.89 & 0.98 & 0.74 & \textbf{0.5} & \textbf{0.59} & 0.24 & 0.65\\
        HetSAGE+clip+add+0.2 & 0.88 & 0.98 & 0.75 & 0.34 & 0.4 & 0.28 & 0.72\\
        HetSAGE+clip+add-only+0.2 & 0.89 & 0.98 & 0.73 & 0.24 & 0.27 & 0.30 & 0.73 \\
        \midrule
        SAGE+clip+0.5 & \textbf{0.86} & \textbf{0.98} & \textbf{0.81} & 0.03 & 0.05 & \textbf{0.33} & \textbf{0.76}\\
        HetSAGE+clip+0.5 & 0.8 & 0.96 & 0.73 & 0.19 & 0.23 & 0.27 & 0.70\\
        HGT+clip+0.5 & 0.81 & 0.96 & 0.73 & \textbf{0.37} & \textbf{0.44} & 0.23 & 0.64\\
        HetSAGE+clip+add+0.5 & 0.79 & 0.96 & 0.72 & 0.22 & 0.25 & 0.29 & 0.70 \\
        HetSAGE+clip+add-only+0.5 & 0.76 & 0.95 & 0.64 & 0.18 & 0.21 & 0.30 & 0.71 \\
    \bottomrule
    \end{tabular}
    \label{tab:corrupt}
\end{table*}

\subsection{Integration of additional layers}

To evaluate the flexibility of our heterogeneous structure, we introduce an additional edge type between global nodes, providing information about the centrality of objects within a scene, derived from geometric data.
This layer is computed by performing geometric collision checks between objects in a scene using the FCL library \cite{pan2012fcl}. Based on the amount of overlap between pairs of colliding objects, a hierarchy is derived using the harmonic centrality metric \cite{boldi2014axioms}. This hierarchy is then integrated into the heterogeneous model as a new edge type between global nodes, providing a topologically different subgraph.

Since GraphSAGE and HGT do not natively support edge features, we implement a specialized edge layer for this edge type, which updates the target node using only the edge features:
\begin{equation}
    x_i' = \gamma(x_i) + \sum_{j \in \mathcal{N}(i)} \phi(x_{ji}),
\end{equation}
where $\mathcal{N}(i)$ denotes the neighbors of node $i$, $x_i'$ is the updated node feature, $x_i$ is the original target node feature, $x_{ji}$ is the edge feature, and $\gamma$ and $\phi$ are learnable linear transformations. The edge feature $x_{ji}$ consists of the amount of geometric overlap between nodes, their respective bounding boxes, and the difference in harmonic centrality as described above.
For all other edge types, GraphSAGE message passing without edge features is used.

We evaluate this new layer both as an additional edge type between global nodes and as a substitute for the original edges resulting from the integration of local SSGs into the global SSG.

Results (see tab. \ref{tab:results} and \ref{tab:corrupt}) show that adding this additional information between global nodes yields performance comparable to HetSAGE+CLIP, with a slight improvement on previously unseen nodes and relationship prediction when using the additional edge type together with the integrated global edges. The same tendency is seen in the evaluation with falsified labels.

Although the improvements in the reported metrics are relatively modest, the results demonstrate that the proposed model can seamlessly integrate multi-modal information into the message passing process without requiring external modules.

\section{Conclusion}

We present a heterogeneous graph model that enables the integration of multi-modal information for incremental 3D scene graph generation. Specifically, we connect information from sensor data frames to a concise global scene graph model built from previous observations. We show that the integration of these prior observations benefits the overall prediction performance on both homogeneous and heterogeneous GNN architectures.

The proposed model demonstrates strong predictive performance for heterogeneous GNN architectures. The integration of additional information sources yields comparable results, indicating that the model effectively incorporates multi-modal data. While homogeneous GNNs achieve high performance on straightforward classification tasks, heterogeneous GNNs are more suited to capture the heterogeneity of multi-modal, semantic information. Furthermore, the heterogeneous graph learning framework offers flexibility for incorporating task-specific information or external knowledge graphs without altering the core architecture.

Future work will explore applying this architecture to full-scale 3D semantic mapping for real-world robotics tasks, integrating additional prior knowledge sources to enhance inference and support explainability.



\bibliographystyle{elsarticle-num}
\bibliography{icmla25}

\begin{thebibliography}{10}
\expandafter\ifx\csname url\endcsname\relax
  \def\url#1{\texttt{#1}}\fi
\expandafter\ifx\csname urlprefix\endcsname\relax\def\urlprefix{URL }\fi
\expandafter\ifx\csname href\endcsname\relax
  \def\href#1#2{#2} \def\path#1{#1}\fi

\bibitem{Armeni2019-ip}
I.~Armeni, Z.-Y. He, J.~Gwak, A.~R. Zamir, M.~Fischer, J.~Malik, S.~Savarese,
  {3D} scene graph: A structure for unified semantics, {3D} space, and camera,
  in: Proceedings of the IEEE International Conference on Computer Vision,
  2019, pp. 5664--5673.

\bibitem{Hughes2022-fg}
N.~Hughes, Y.~Chang, L.~Carlone, Hydra: A real-time spatial perception system
  for {3D} scene graph construction and optimization, in: Robotics: Science and
  Systems XVIII, Robotics: Science and Systems Foundation, 2022.
\newblock \href {https://doi.org/10.15607/rss.2022.xviii.050}
  {\path{doi:10.15607/rss.2022.xviii.050}}.

\bibitem{Bavle2023-iz}
H.~Bavle, J.~L. Sanchez-Lopez, M.~Shaheer, J.~Civera, H.~Voos,
  \textit{S-graphs+:} real-time localization and mapping leveraging
  hierarchical representations, IEEE Robot. Autom. Lett. 8~(8) (2023)
  4927--4934.
\newblock \href {https://doi.org/10.1109/lra.2023.3290512}
  {\path{doi:10.1109/lra.2023.3290512}}.

\bibitem{Looper2023-kb}
S.~Looper, J.~Rodriguez-Puigvert, R.~Siegwart, C.~Cadena, L.~Schmid, {3D}
  {VSG}: Long-term semantic scene change prediction through {3D} variable scene
  graphs, in: 2023 IEEE International Conference on Robotics and Automation
  (ICRA), IEEE, 2023, pp. 8179--8186.
\newblock \href {https://doi.org/10.1109/ICRA48891.2023.10161212}
  {\path{doi:10.1109/ICRA48891.2023.10161212}}.

\bibitem{Ravichandran2022-vd}
Z.~Ravichandran, L.~Peng, N.~Hughes, J.~D. Griffith, L.~Carlone, Hierarchical
  representations and explicit memory: Learning effective navigation policies
  on {3D} scene graphs using graph neural networks, Proceedings - IEEE
  International Conference on Robotics and Automation (2022) 9272--9279\href
  {https://doi.org/10.1109/ICRA46639.2022.9812179}
  {\path{doi:10.1109/ICRA46639.2022.9812179}}.

\bibitem{Li2022-ni}
X.~Li, D.~Guo, H.~Liu, F.~Sun, Embodied semantic scene graph generation, in:
  A.~Faust, D.~Hsu, G.~Neumann (Eds.), Proceedings of the 5th Conference on
  Robot Learning, Vol. 164 of Proceedings of Machine Learning Research, PMLR,
  2022, pp. 1585--1594.

\bibitem{Agia2022-mi}
C.~Agia, K.~M. Jatavallabhula, M.~Khodeir, O.~Miksik, V.~Vineet, M.~Mukadam,
  L.~Paull, F.~Shkurti, Taskography: Evaluating robot task planning over large
  {3D} scene graphs, in: A.~Faust, D.~Hsu, G.~Neumann (Eds.), Proceedings of
  the 5th Conference on Robot Learning, Vol. 164 of Proceedings of Machine
  Learning Research, PMLR, 2022, pp. 46--58.

\bibitem{Wald2020-yj}
J.~Wald, H.~Dhamo, N.~Navab, F.~Tombari, Learning {3D} semantic scene graphs
  from {3D} indoor reconstructions, in: 2020 IEEE/CVF Conference on Computer
  Vision and Pattern Recognition (CVPR), IEEE, 2020.
\newblock \href {https://doi.org/10.1109/cvpr42600.2020.00402}
  {\path{doi:10.1109/cvpr42600.2020.00402}}.

\bibitem{Wu2021-hw}
S.-C. Wu, J.~Wald, K.~Tateno, N.~Navab, F.~Tombari, {SceneGraphFusion}:
  Incremental {3D} scene graph prediction from {RGB}-{D} sequences, in: 2021
  IEEE/CVF Conference on Computer Vision and Pattern Recognition (CVPR), IEEE,
  2021.
\newblock \href {https://doi.org/10.1109/cvpr46437.2021.00743}
  {\path{doi:10.1109/cvpr46437.2021.00743}}.

\bibitem{Wu2023-dm}
S.-C. Wu, K.~Tateno, N.~Navab, F.~Tombari, Incremental {3D} semantic scene
  graph prediction from {RGB} sequences, Proc. IEEE Comput. Soc. Conf. Comput.
  Vis. Pattern Recognit. (2023) 5064--5074\href
  {https://doi.org/10.1109/CVPR52729.2023.00490}
  {\path{doi:10.1109/CVPR52729.2023.00490}}.

\bibitem{Feng2025-cz}
M.~Feng, C.~Yan, Z.~Wu, W.~Dong, Y.~Wang, A.~Mian, History-enhanced {3D} scene
  graph reasoning from {RGB}-{D} sequences, IEEE Transactions on Circuits and
  Systems for Video Technology (2025) 1--1\href
  {https://doi.org/10.1109/TCSVT.2025.3548308}
  {\path{doi:10.1109/TCSVT.2025.3548308}}.

\bibitem{Ma2024-gd}
Y.~Ma, H.~Liu, Y.~Pei, Y.~Guo, Heterogeneous graph learning for scene graph
  prediction in {3D} point clouds, ECCV (2024) 274--291\href
  {https://doi.org/10.1007/978-3-031-73347-5\_16}
  {\path{doi:10.1007/978-3-031-73347-5\_16}}.

\bibitem{Giuliari2023-sw}
F.~Giuliari, G.~Skenderi, M.~Cristani, A.~D. Bue, Y.~Wang, Leveraging
  commonsense for object localisation in partial scenes, IEEE Trans. Pattern
  Anal. Mach. Intell. 45~(10) (2023) 12038--12049.
\newblock \href {https://doi.org/10.1109/TPAMI.2023.3272523}
  {\path{doi:10.1109/TPAMI.2023.3272523}}.

\bibitem{Speer2017-os}
R.~Speer, J.~Chin, C.~Havasi, {ConceptNet} 5.5: An open multilingual graph of
  general knowledge, Proc. Conf. AAAI Artif. Intell. 31~(1) (12~Feb. 2017).
\newblock \href {https://doi.org/10.1609/AAAI.V31I1.11164}
  {\path{doi:10.1609/AAAI.V31I1.11164}}.

\bibitem{Qiu2023-ou}
Y.~Qiu, H.~I. Christensen, {3D} scene graph prediction on point clouds using
  knowledge graphs, IEEE International Conference on Automation Science and
  Engineering 2023-August (2023).
\newblock \href {https://doi.org/10.1109/CASE56687.2023.10260650}
  {\path{doi:10.1109/CASE56687.2023.10260650}}.

\bibitem{Zareian2020-kc}
A.~Zareian, S.~Karaman, S.-F. Chang, Bridging knowledge graphs to generate
  scene graphs, in: Computer Vision -- ECCV 2020: 16th European Conference,
  Glasgow, UK, August 23--28, 2020, Proceedings, Part XXIII, Springer-Verlag,
  Berlin, Heidelberg, 2020, pp. 606--623.
\newblock \href {https://doi.org/10.1007/978-3-030-58592-1\_36}
  {\path{doi:10.1007/978-3-030-58592-1\_36}}.

\bibitem{Wald2019-ki}
J.~Wald, A.~Avetisyan, N.~Navab, F.~Tombari, M.~Niessner, {RIO}: {3D} object
  instance re-localization in changing indoor environments, in: 2019 IEEE/CVF
  International Conference on Computer Vision (ICCV), IEEE, 2019.
\newblock \href {https://doi.org/10.1109/iccv.2019.00775}
  {\path{doi:10.1109/iccv.2019.00775}}.

\bibitem{Zhang2021-tl}
C.~Zhang, J.~Yu, Y.~Song, W.~Cai, Exploiting edge-oriented reasoning for {3D}
  point-based scene graph analysis, in: 2021 IEEE/CVF Conference on Computer
  Vision and Pattern Recognition (CVPR), IEEE, 2021, pp. 9700--9710.
\newblock \href {https://doi.org/10.1109/CVPR46437.2021.00958}
  {\path{doi:10.1109/CVPR46437.2021.00958}}.

\bibitem{radford2021learning}
A.~Radford, J.~W. Kim, C.~Hallacy, A.~Ramesh, G.~Goh, S.~Agarwal, G.~Sastry,
  A.~Askell, P.~Mishkin, J.~Clark, et~al., Learning transferable visual models
  from natural language supervision, in: International conference on machine
  learning, PmLR, 2021, pp. 8748--8763.

\bibitem{Qi2017-wl}
C.~R. Qi, H.~Su, K.~Mo, L.~J. Guibas, {PointNet}: Deep learning on point sets
  for {3D} classification and segmentation, in: Proceedings of the IEEE
  Conference on Computer Vision and Pattern Recognition (CVPR), 2017.

\bibitem{Hamilton2017-nc}
W.~L. Hamilton, R.~Ying, J.~Leskovec, Inductive representation learning on
  large graphs, Adv. Neural Inf. Process. Syst. 2017-December (2017)
  1025--1035.

\bibitem{Schlichtkrull2018-fx}
M.~Schlichtkrull, T.~N. Kipf, P.~Bloem, R.~Van Den~Berg, I.~Titov, M.~Welling,
  Modeling relational data with graph convolutional networks, in: Proc.
  European Semantic Web Conference (ESWC), 2018, pp. 593--607.

\bibitem{Hu2020-gs}
Z.~Hu, Y.~Dong, K.~Wang, Y.~Sun, Heterogeneous graph transformer, in:
  Proceedings of The Web Conference 2020, WWW '20, Association for Computing
  Machinery, New York, NY, USA, 2020, pp. 2704--2710.
\newblock \href {https://doi.org/10.1145/3366423.3380027}
  {\path{doi:10.1145/3366423.3380027}}.

\bibitem{Gkanatsios2019-eo}
N.~Gkanatsios, V.~Pitsikalis, P.~Koutras, P.~Maragos,
  Attention-translation-relation network for scalable scene graph generation,
  in: 2019 IEEE/CVF International Conference on Computer Vision Workshop
  (ICCVW), IEEE, 2019.
\newblock \href {https://doi.org/10.1109/iccvw.2019.00218}
  {\path{doi:10.1109/iccvw.2019.00218}}.

\bibitem{pan2012fcl}
J.~Pan, S.~Chitta, D.~Manocha, Fcl: A general purpose library for collision and
  proximity queries, in: 2012 IEEE international conference on robotics and
  automation, IEEE, 2012, pp. 3859--3866.

\bibitem{boldi2014axioms}
P.~Boldi, S.~Vigna, Axioms for centrality, Internet Mathematics 10~(3-4) (2014)
  222--262.

\end{thebibliography}

\end{document}